\pdfoutput=1

\documentclass[11pt]{article}

\usepackage{emnlp2021}
\usepackage{times}
\usepackage[utf8]{inputenc}
\usepackage{latexsym}
\usepackage{graphicx}
\usepackage[T1, T2A]{fontenc}
\usepackage[T1]{fontenc}
\usepackage[russian,english]{babel}
\usepackage[normalem]{ulem}
\usepackage{float}
\usepackage{multirow}

\useunder{\uline}{\ul}{}

\usepackage{microtype}

\title{\textit{Don't Go Far Off}: An Empirical Study on Neural Poetry Translation}

\author{Tuhin Chakrabarty \textsuperscript{1},
  Arkadiy Saakyan\textsuperscript{1}, 
  \textbf{and} \textbf{Smaranda Muresan}\textsuperscript{1,2}\\ 
  \textsuperscript{1}Department of Computer Science, Columbia University \\
  \textsuperscript{2}Data Science Institute, Columbia University\\
  {\tt tuhin.chakr@cs.columbia.edu}, {\tt \{a.saakyan,smara\}@columbia.edu}
  }

\begin{document}
\maketitle
\begin{abstract}

Despite constant improvements in machine translation quality, automatic poetry translation remains a challenging problem due to the lack of open-sourced parallel poetic corpora, and to the intrinsic complexities involved in preserving the semantics, style and figurative nature of poetry. We present an empirical investigation for poetry translation along several dimensions: 1) size and style of training data (poetic vs. non-poetic), including a zero-shot setup; 2) bilingual vs. multilingual learning; and 3) language-family-specific models vs. mixed-language-family models. To accomplish this, we contribute a parallel dataset of poetry translations for several language pairs. Our results show that \emph{multilingual} fine-tuning on \emph{poetic} text significantly outperforms \emph{multilingual} fine-tuning on \emph{non-poetic} text that is 35X larger in size, both in terms of automatic metrics (BLEU, BERTScore, COMET) and human evaluation metrics such as faithfulness (meaning and poetic style). Moreover, \emph{multilingual} fine-tuning on poetic data outperforms \emph{bilingual} fine-tuning on poetic data.
\footnote{The italics part of the title is the translation of a poem by Pablo Neruda with the same name.} 
\end{abstract}

\section{Introduction}

American poet Robert Frost once defined poetry as \textit{``that which gets lost out of both prose and verse in translation”}  \cite{frost1961conversations}. Indeed, the task is so complex that translators often have to ``create a poem in the target language which is readable and enjoyable as an independent, literary text" \cite{jones2011translation}. But even though poetry is destined to lose its accuracy, integrity, and beauty even in human translation, the process conceives new opportunities to stress-test the ability of machine translation models to deal with figurative language. 

\begin{table}
\small
\centering
\begin{tabular}{|l|l|}
\hline
Original   & \begin{tabular}[c]{@{}l@{}}Il ny avait que sable et boue\\ Où sétait ouverte la tombe.\\ Le long des murs de la prison\\ On ne voyait aucune tombe\end{tabular}   \\ \hline\hline
Human      & \begin{tabular}[c]{@{}l@{}}For where a grave had opened wide,\\ There was no grave at all:\\ Only a stretch of mud and sand\\ By the hideous prison-wall,\end{tabular} \\ \hline\hline
GCK & \begin{tabular}[c]{@{}l@{}}But there was only sand and mud.\\ To where the grave was laid.\\ Along the walls of prison wall.\\ We saw no \textbf{\color{red}masquerade}\end{tabular}       \\ \hline\hline
Google     & \begin{tabular}[c]{@{}l@{}}There was only sand and mud\\ Where the grave had opened.\\ Along the prison walls\\ No tomb could be seen.\end{tabular}                  \\ \hline
\end{tabular}
\caption{\label{Table:1} A French poem accompanied by human translation,  \cite{ghazvininejad2018neural} (GCK) system translation, and Google Translate.\footnotemark{}}
\end{table}
\footnotetext{Example taken from \cite{ghazvininejad2018neural}.}
\begin{table*}
\centering
\small
\begin{tabular}{|l|l|c|c|c|}
\hline
Language Pair                   & Source                                 &  Train &  Valid  &  Test    \\ \hline
\multirow{2}{*}{Spanish-English}   & \url{https://www.poesi.as/}                 & \multirow{2}{*}{37,746} & \multirow{2}{*}{2059} & \multirow{2}{*}{536} \\ \cline{2-2}
                           & \url{https://lyricstranslate.com/}          &  &                      &   \\ \hline\hline
\multirow{1}{*}{Russian-English}                    & \url{https://ruverses.com/}                 & 50,001   &  4186            &  548 \\ \hline\hline
\multirow{3}{*}{Portugese-English} & \url{http://www.poemsfromtheportuguese.org/} & \multirow{3}{*}{15,199} & \multirow{3}{*}{699} & \multirow{3}{*}{140} \\ \cline{2-2}
                           & \url{https://www.poetryinternational.org/}   &             &          &  \\ \cline{2-2}
                           & \url{https://lyricstranslate.com/}           &           &            &  \\ \hline\hline
\multirow{3}{*}{German-English}    & \url{http://www.poemswithoutfrontiers.com/}  & \multirow{3}{*}{17,000} & \multirow{3}{*}{1,050}& \multirow{3}{*}{1295} \\ \cline{2-2}
                           & \url{https://www.poetryinternational.org/}   &              &         &  \\ \cline{2-2}
                           & \url{https://lyricstranslate.com/}           &             &          &  \\ \hline\hline
\multirow{3}{*}{Italian-English}    & \url{https://digitaldante.columbia.edu/}  & \multirow{3}{*}{34,534} & \multirow{3}{*}{1,997} & \multirow{3}{*}{528} \\ \cline{2-2}
                           & \url{https://www.poetryinternational.org/}   &     &                  &  \\ \cline{2-2}
                           & \url{https://lyricstranslate.com/}           &             &         &   \\ \hline\hline
\multirow{2}{*}{Dutch-English}   & \url{https://www.poetryinternational.org/}                 & \multirow{2}{*}{23,403}  & \multirow{2}{*}{1,000} &  \multirow{2}{*}{159}\\ \cline{2-2}
                           & \url{https://lyricstranslate.com/}          &              &         &  \\ \hline                           
\end{tabular}
\caption{\label{table:datastat}Dataset source and statistics}
\end{table*}

While most computational work has focused on poetry generation \cite{ hopkins-kiela-2017-automatically,uthus2021augmenting,van2020automatic,ghazvininejad2016generating, li-etal-2020-rigid, hamalainen-alnajjar-2019-generating, yi2020mixpoet, deng2019iterative, ijcai2019-684,yang-etal-2018-stylistic}, research on poetry translation is in its infancy 
\cite{ghazvininejad2018neural,genzel-etal-2010-poetic}. 

For example, \citet{ghazvininejad2018neural} employs a  constrained decoding technique to maintain rhyme in French to English poetry translation. However, while keeping the poetic style and fluency, the translation might diverge in terms of meaning w.r.t. the input. Table \ref{Table:1} shows how the system generates a semantically inconsistent word ``masquerade" to rhyme with ``laid", whereas the original poem talks about ``tomb". 

Meanwhile, state-of-the-art machine translation systems trained on large non-poetic data might preserve meaning and fluency, but not the poetic style (e.g., Google Translate's output in Table \ref{Table:1}).  

Two main challenges exist for automatic poetry translation: the lack of open-sourced multilingual parallel poetic corpora and the intrinsic complexities involved in preserving the semantics, style and figurative nature of poetry. To address the first, we collect a multilingual parallel corpus consisting of more than 190,000 lines of poetry spanning over six languages. We try to tackle the second challenge by leveraging multilingual pre-training (e.g., mBART \cite{liu2020multilingual}) and multilingual fine-tuning \cite{tang2020multilingual,aharoni-etal-2019-massively} that have recently led to advances in neural machine translation for low-resource languages. Moreover, it has been shown that adaptive pre-training and/or fine-tuning on in-domain data always lead to improved performance on the end task \cite{gururangan-etal-2020-dont}. 

Since poetry translation falls into the low-resource (no or little parallel data) and in-domain translation scenarios, we present an \textit{empirical investigation} on whether advances in these areas bring us a step closer to poetry translation systems that \emph{don't go far off} in terms of faithfulness (i.e., keeping the meaning and poetic style of the input).  

We make the following contributions:
\begin{itemize}
    \item We release several parallel poetic corpora enabling translation from Russian, Spanish, Italian,  Dutch, German, and Portuguese to English. We also release test sets for poetry translation from Romanian, Ukranian and Swedish to evaluate the zero-shot performance of our models. 
    \item We show that multilingual fine-tuning on poetic text significantly outperforms multilingual fine-tuning on non-poetic text that is 35X larger in size (177K vs 6M), both in terms of automatic and human evaluation metrics such as faithfulness. However, for the bilingual case the pattern is not so evident. Moreover, multilingual fine-tuning on poetic data outperforms bilingual fine-tuning on poetic data. The latter two results showcase the importance of multilingual fine-tuning for poetry translation. 
    \item We also show that multilingual fine-tuning on languages belonging to the same language family sometimes leads to improvement over fine-tuning on all languages.
\end{itemize}

\begin{table*}
\small
\begin{tabular}{|p{7.45cm}|l|}
\hline
Russian & English                                                                                                    \\ \hline\begin{tabular}[c]{@{}l@{}}
\begin{otherlanguage*}{russian}Они любили друг друга так долго и нежно,\end{otherlanguage*}\\\begin{otherlanguage*}{russian}С тоской глубокой и страстью безумно-мятежной!\end{otherlanguage*}\\  \end{tabular}
\begin{tabular}[c]{@{}l@{}}\end{tabular}                     & \begin{tabular}[c]{@{}l@{}}\textit{Their love was so gentle, so long, and surprising,}\\ \textit{With pining, so deep, and zeal, like a crazy uprising!}\end{tabular} \\ \hline
Spanish & English                                                                                                                                                                                                                                                                   \\ \hline
\begin{tabular}[c]{@{}l@{}}Puedo escribir los versos más tristes esta noche.\\ Yo la quise, y a veces ella también me quiso.\\\end{tabular} & \begin{tabular}[c]{@{}l@{}}\textit{I can write the saddest lines tonight.}\\ \textit{I loved her, sometimes she loved me too.}\\ \end{tabular}                                                    \\ \hline
Portugese & English                                                                                                                                                                                                                                                                   \\ \hline
\begin{tabular}[c]{@{}l@{}}Num jardim adornado de verdura\\ a que esmaltam por cima várias flores\\\end{tabular} & \begin{tabular}[c]{@{}l@{}}\textit{To a garden luxuriously verdant
}\\ \textit{and enamelled with countless flowers}\\ \end{tabular}                                                    \\ \hline
German & English                                                                                                                                                                                                                                                                   \\ \hline

\begin{tabular}[c]{@{}l@{}}wir opfern zuerst deine keuschheit, liebster\\ und erhalten die gabe der sprache dafür\\\end{tabular} & \begin{tabular}[c]{@{}l@{}}\textit{we'll sacrifice your chastity first, dearest
}\\ \textit{and get the gift of language in return
}\\\end{tabular}                                                    \\ \hline
Italian & English                                                                                                                                                                                                                                                                   \\ \hline

\begin{tabular}[c]{@{}l@{}}Tonda, gelida dei suoi oceani, trasparente\\ come una cellula sotto il microscopio\\\end{tabular} & \begin{tabular}[c]{@{}l@{}}\textit{Round, frozen in its oceans, transparent}\\ \textit{like a cell under the microscope}\\ \end{tabular}                                                    \\ \hline
Dutch & English                                                                                                                                                                                                                                                                   \\ \hline

\begin{tabular}[c]{@{}l@{}}Avond en het breeklicht in je ogen en je kijkt.\\ het breekt oranje op in je ogen het vloeiende licht\\\end{tabular} & \begin{tabular}[c]{@{}l@{}}\textit{Evening and the glow stick’s in your eyes and you are looking}\\ \textit{its orange snapped into your eyes the liquid light}\\\end{tabular}                                                    \\ \hline
\end{tabular}
\caption{\label{table2} Parallel Poetic translations written by humans from our multilingual datasets.}
\end{table*}

Beyond advancing poetic translation, our findings will be helpful for other figurative language or literary text translation tasks. Our code and data and can be found in \url{https://github.com/tuhinjubcse/PoetryTranslationEMNLP2021} while our pre-trained models can be found at \url{https://huggingface.co/TuhinColumbia}. We hope that the data, models and the code released will encourage further research in this area.

\section{Datasets}  \label{poetic_data}

\subsection{Poetic Training Data}\label{ref:poetic}

Given the lack of available multilingual poetic corpora, we collect several medium-scale parallel datasets. We identify websites that provide English translations for Spanish (Es), Russian (Ru), Portuguese (Pt), German (De), Italian (It) and Dutch (Nl) poetry. Table \ref{table:datastat} shows the number of parallel sentences for each language pair as well as the websites from which they have been collected. Given that most of the websites were specifically designed for poetry translation, where translations are typically written by experts (professional translators), we believe our data to be of high quality. We make a simplifying assumption and focus on line-by-line translation. Thus, during scraping from these websites, we discard translations that are different in the number of lines from the original poems. 
We collect approximately 190K (with 177K in training) parallel poetic lines spanning 6 different languages (see Table \ref{table2} for examples). This data is further split into train and validation.

\subsection{Non-Poetic Training Data} \label{ref:nonpoetic}
We also benchmark the quality of poetry translations obtained by models trained on non-poetic data. For this we rely on OPUS100 corpus \cite{tiedemann-2012-parallel} as well as the ML50 corpus\cite{tang2020multilingual} designed to demonstrate the impact of multilingual fine-tuning. Each of the language pairs in OPUS100 have 1 million parallel sentences in their training set, several orders of magnitude larger than our poetic parallel data. For example, Portuguese-English non-poetic data is $65$ times larger than the poetic data, while the Russian-English  non-poetic data is $18$ times larger than the poetic data. The size of the smallest non-poetic parallel corpus is about $6$ times larger than all our poetic parallel data combined. For ML50  \cite{tang2020multilingual}, benchmark data is collected across 50 languages from publicly available datasets such as WMT, IWSLT, WAT, TED. The size of 
parallel sentences in ML50 corresponding to the languages under study are: De (45.8M), Es (14.5M), Ru (13.9M), Nl (0.23M), It (0.2M), and Pt (0.04M).

\subsection{Test Data}
We create a high quality blind test set for every language independent of data mentioned in Table \ref{table:datastat} by carefully hand-picking poems unseen in the training or validation set. Every line has a single reference. Our blind test consists of $3522$ sentences spanning across $209$ poems in $6$ languages. Our combined multilingual test set consists of $548$ lines in Russian, $536$ lines in Spanish, $528$ lines in Italian, $1295$ lines in German, $140$ lines in Portuguese and $159$ lines in Dutch. We also test our models on $7$ Ukrainian poems ($100$ lines), $8$ Romanian poems ($100$ lines) and $7$ Swedish poems ($100$ lines) in a zero-shot setting.  

\section{Methods}

mBART \cite{liu2020multilingual} is a multilingual sequence-to-sequence (seq2seq) denoising auto-encoder, which 
is trained by applying the BART objective \cite{lewis2019bart} to large-scale monolingual corpora across many languages. The input texts are noised by masking phrases and permuting sentences, and a single Transformer model is learned to recover the texts. 
Unlike other pre-training approaches for machine translation, mBART pre-trains a complete autoregressive seq2seq model. It is trained once for all languages, providing a set of parameters that can be fine-tuned for any of the language pairs for supervised machine translation without any task-specific or language-specific modifications or initialization schemes. 
For supervised sentence-level MT, mBART initialization leads to significant gains (up to $12$ BLEU points) across low/medium-resource pairs ($<10$M bi-text pairs). This makes mBART an ideal candidate for our task of poetry translation given the scale of our parallel corpora.

However, while mBART was trained on a variety of languages, the multilingual nature of the pre-training is not used during fine-tuning. To solve this, \citet{tang2020multilingual} propose \emph{multilingual fine-tuning} of pre-trained models, and demonstrate large improvements compared to bilingual fine-tuning. 
They explore 3 configurations to create different versions of multilingual translation models: Many-to-one (N $\rightarrow$ 1), one-to-Many (1 $\rightarrow$ N), and Many-to-Many (N$ \leftrightarrow$N) via a pivot language. The \textit{Many-to-one} model encodes N languages and decodes to English. Given that we are translating poems in various languages to English, we further fine-tune the \textit{Many-to-one} model for our task.

\subsection{Implementation Details}
For bilingual fine-tuning on poetic data, we use the \textit{mbart-large-50} checkpoint from \cite{wolf-etal-2020-transformers}, and fine-tune it for up to $8$ epochs, saving the best checkpoint based on eval-BLEU scores. For bilingual fine-tuning on non-poetic data, we fine-tune the model for $3$ epochs. For multilingual fine-tuning, we use the \textit{mbart-large-50-many-to-one-mmt}. We perform multilingual fine-tuning for $3$ epochs for both poetic/non-poetic data. We use the same hyperparameters as the standard huggingface implementation. We use ($2$-$4$) nvidia A100 GPUs for fine-tuning pretrained checkpoints. For fine-tuning mBART on non-poetic data, we set the \textit{gradient\_accumulation\_steps} to $10$ and batch size to $8$ while for poetic fine-tuning we vary batch size between $24$ and $32$, and set \textit{gradient\_accumulation\_steps} to $1$.

To perform multilingual fine-tuning, we concatenate bitexts of different language pairs (i, j) into a collection $B_{i,j} = {(x_{i}, y_{j} )}$ for each direction (i, j). Following mBART \cite{liu2020multilingual}, we augment each bitext  $(x_{i}, y_{j} )$ by adding a source and a target language token at the beginning of $x$ and $y$, respectively, to form a target language token augmented pair $(x_{0}, y_{0})$. We then initialize transformer based seq-to-seq model by the pretained mBART, and provide the multilingual bitexts $B = \cup_{i,j} B_{i,j}$ to fine-tune the pretrained model.

\section{Experimental Setting}

We experiment with several systems to evaluate performance across several dimensions: poetic vs non-poetic data; multilingual fine-tuning vs. bilingual fine-tuning; language-family-specific models vs. mixed-language-family models.

\begin{itemize}
    \item \textbf{Non-Poetic Bi (OPUS)}: fine-tuned mBART50 on Non-Poetic data from OPUS100 (Section \ref{ref:nonpoetic}) for respective languages bilingually.
    \item \textbf{Non-Poetic Multi (ML50)}:  mBART-large-50-many-to-one model implemented in  the huggingface package. 
    This is a multilingually fine-tuned model on 50 languages from the ML50 data that is 
    4 times larger than OPUS and created using all of the data that is publicly available (e.g., WMT, IWSLT, WAT, TED).
    \item \textbf{Non-Poetic Multi (OPUS)}: multilingually fine-tuned mBART-large-50-many-to-one model on Non-Poetic data for 6 languages from OPUS100 (Section \ref{ref:nonpoetic}) (6M parallel sentences).
    \item  \textbf{Poetic}: fine-tuned mBART50 bilingually (e.g., Ru-En, Es-En, It-En) on poetic data described in Section  \ref{ref:poetic}. 
    \item  \textbf{Poetic All}: multilingually fine-tuned mBART-large-50-many-to-one  on all poetic data combined.
    \item  \textbf{Poetic LangFamily}: multilingually  fine-tuned mBART-large-50-many-to-one on poetic data for all languages belonging to the same language family. For instance, Pt, Es, It belong to the Romance language family, while De and Nl are both Germanic languages.
\end{itemize}

\subsection{Automatic Evaluation Setup}
For the automatic evaluation, we compare the performance of all the above mentioned models in terms of three metrics: BLEU, BERTScore and COMET. 

\textit{BLEU}~\cite{BLEU} is one of the most widely used automatic evaluation metrics for  Machine Translation. We use the SacreBLEU \cite{post-2018-call} python library to compute BLEU scores between the system output and the human written gold reference. 

\textit{BERTScore}~\cite{zhang2019bertscore} has been used recently for evaluating text generation systems using contextualized embeddings, and it is said to somewhat ameliorate the problems with BLEU. BERTScore also has better correlation with human judgements \cite{zhang2019bertscore}.
It computes a similarity score using contextual embeddings for each token in the system output with each token in the reference. We report F1-Score of \textit{BERTScore}. We use the latest  implementation to date which replaces BERT with \textit{deberta-large-mnli}, which is a DeBERTa model \cite{he2020deberta} fine-tuned on MNLI \cite{williams2017broad}.

Recently \citet{kocmi2021ship} criticized the use of BLEU through a systematic study of 4380 machine translation systems and recommend use of a pre-trained metric COMET \cite{rei-etal-2020-comet}. COMET leverages recent breakthroughs in cross-lingual pre-trained language modeling resulting in highly multilingual and adaptable MT evaluation models that exploit information from both the source input and a target-language reference translation in order to more accurately predict MT quality. We rely on the recommended model \texttt{wmt-large-da-estimator-1719}, which is trained to minimize the mean squared error between the predicted scores and the DA \cite{graham-etal-2013-continuous} quality assessments. Notice that these scores are normalized per annotator and hence not bounded between 0 and 1, allowing negative scores to occur; higher score means better translation.

\subsection{Human-based Evaluation Setup}
Even though arguably useful for evaluating meaning preservation, automatic metrics are not as suitable to measure other aspects of poetic translation such as the use of figurative language and style. We conduct human evaluation by recruiting three bilingual speakers as volunteers for each language. NMT systems are susceptible to producing highly pathological translations that are completely unrelated to the source input often termed as \textit{hallucinations} \cite{raunak-etal-2021-curious}(e.g., the word \textit{Lungs} in Table \ref{hallucination}). To account for these effects, we use \textit{faithfulness} as a measure that combines both \textit{meaning preservation and poetic style}. 

We evaluate the best translations from multilingual models trained on poetic and non-poetic data. Human judges were asked to evaluate on a binary scale whether: i) the model introduces hallucinations or translates the input into something arbitrary, i.e. (\textit{Are they keeping the meaning of the input text?}) and at the same time ii) the syntactic structure is poetic and the translations are rich in poetic figures of speech (e.g., metaphors, similes, personification).

In this evaluation we compare the multilingually fine-tuned models on Non-Poetic data (Non-Poetic Multi (OPUS) and Non-Poetic Multi (ML50)) vs. multilingually fined-tuned models on Poetic data (Poetic All and Poetic LangFamily). We chose the best model in each category based on the BERTScore in the automatic evaluation. 

We chose a subset of the test set for human evaluation: 1044 sentences spanning across 80 poems in 6 languages (204 lines in Russian, 173 lines in Italian, 140 lines in Portuguese, 220 lines in Spanish, 148 lines in German, and 159 lines in Dutch with corresponding human translations). Human judges were also provided with gold translations to make the judgement easier.
Agreement rates were measured using Krippendorff's $\alpha$ and a moderate agreement of 0.61 was achieved.

\section{Results}

Our results based on automatic metrics are summarized in  Table \ref{results} and the human evaluation in Table \ref{human}. The first insight is that multilingual fine-tuning on Poetic data (Poetic All and Poetic LangFamily) outperforms mutilingual fine-tuning on Non-Poetic data (Non-Poetic Multi (ML50, Opus)) for all languages both in terms of automatic metrics (BLEU and BERTScore) and human evaluation based on faithfulness (Table \ref{human}).  
Between Poetic-All and Non-Poetic Multi we see at least $2.5$ point improvement in BLEU scores as well as $1$ point improvement in BertScore in translation of every language pair. For the recently developed metric COMET, we see that the best models are the multilingually fine-tuned poetic models, which is consistent with the results obtained using the other two metrics.

\begin{table*}[t]
\small
\centering
\begin{tabular}{|l|c|c|c|}
\hline
\textbf{Model}       & \multicolumn{1}{l|}{\textbf{BLEU}} & \multicolumn{1}{l|}{\textbf{BERTScore}} & \multicolumn{1}{l|}{\textbf{COMET}} \\ \hline
Non-Poetic Bi(OPUS) Ru-En            & 12.4                               & 65.4      & -47.83 \\ \hline
Non-Poetic Multi(ML50) Ru-En         &    13.0           &               67.5    &  -37.55 \\ \hline
Non-Poetic Multi(OPUS) Ru-En         &    12.8          &               67.2    & -39.5 \\ \hline

Poetic Ru-En         & 11.9                              & 64.3                 &           -55.14        \\ \hline
Poetic LangFamily & -                      & -             &       -       \\ \hline
Poetic All & \textbf{17.0}                      & \textbf{70.2}                     &   \textbf{-25.71}  \\ \hline\hline
Non-Poetic Bi(OPUS) Es-En            & 26.9                               & 74.6         &  1.43\\ \hline
Non-Poetic Multi(ML50) Es-En         & 5.1               &  58.9                  & -60.98 \\ \hline
Non-Poetic Multi(OPUS) Es-En         & 28.0               &  75.6                &  4.84 \\ \hline
Poetic Es-En         & 26.8                               & 74.3                   &       -3.09          \\ \hline
Poetic LangFamily &    30.9                & \textbf{77.2}                       &  \textbf{12.14 } \\ \hline
Poetic All & \textbf{31.2}                    & 76.6
    &           10.10        \\ \hline\hline
Non-Poetic Bi(OPUS) Pt-En            &    9.5                            &    63.3                           &   -47.27   \\ \hline
Non-Poetic Multi(ML50) Pt-En         & 7.3              &  62.7                 &  -53.48 \\ \hline
Non-Poetic Multi(OPUS) Pt-En         & 9.2              &  64.0                 &  -42.86 \\ \hline
Poetic Pt-En         &  9.6                             &    63.4                            &   -50.93   \\ \hline
Poetic LangFamily &    \textbf{12.5}                 & 66.4     &  -39.36 \\ \hline
Poetic All & 12.2                     & \textbf{66.6}                   &  \textbf{ -35.89 }  \\ \hline\hline

Non-Poetic Bi(OPUS) It-En            &  22.2                              &     70.3     &   -14.85  \\ \hline
Non-Poetic Multi(ML50) It-En         & 17.0               &  68.7      & -24.53 \\ \hline
Non-Poetic Multi(OPUS) It-En         & 22.9               &  71.1                & -8.87  \\ \hline
Poetic It-En         &  18.8                             &    69.3         &  -24.21 \\ \hline
Poetic LangFamily &    \textbf{25.4}                 & \textbf{72.2}               &      \textbf{-7.35 }   \\\hline
Poetic All & 24.6                      &  71.6                        &  -8.87 \\ \hline\hline

Non-Poetic Bi(OPUS) De-En                                           & 15.2  & 68.6     &    -27.95  \\ \hline
Non-Poetic Multi(ML50) De-En         & 20.1               &              73.4   &  -5.88 \\ \hline
Non-Poetic Multi(OPUS) De-En         & 17.8              &              70.9    & -16.77 \\ \hline
Poetic De-En         &  16.8                             &    70.2                        &     -23.07     \\ \hline
Poetic LangFamily &    20.5               &  73.6                   &     -4.22 \\ \hline
Poetic All & \textbf{22.7}                      & \textbf{74.6}     &          \textbf{-0.52 }         \\ \hline\hline

Non-Poetic Bi(OPUS) Nl-En                                     & 24.5  &  72.5       &   -4.83   \\ \hline
Non-Poetic Multi(ML50) Nl-En         &    23.8            &    72.2  & -6.73 \\ \hline
Non-Poetic Multi(OPUS) Nl-En         &    26.1            &    72.9 &  -4.83 \\ \hline
Poetic Nl-En         &    26.5                          &        71.6        &          -12.73           \\ \hline
Poetic LangFamily &     \textbf{32.1}             &                    74.3    & -3.74  \\ \hline
Poetic All &            30.7           &         \textbf{74.5}            &   \textbf{-1.90 }  \\ \hline
\end{tabular}
\caption{\label{results} Performance of mBART fine-tuned on different datasets in terms of automatic evaluation metrics on test data in various settings. Difference is significant, $( \alpha < 0.005)$ via Wilcoxon signed-rank test.}
\end{table*}

However, when comparing the bilingually fine-tuned models  (Poetic vs. Non-Poetic Bi(Opus)) the pattern is not as clear based on automatic metrics. We see comparable performance, but not a clear winner across languages and metrics. However, as with the multilingual case, the size of Poetic data is much smaller than the Non-Poetic data (20X to 50X smaller depending on the language). 
We also mixed poetic and non-poetic data in equal proportion and fine-tuned mBART by framing it as a domain adaption problem, however it did not lead to significant improvements and degenerated in a few languages. We also tried intermediate fine-tuning \cite{phang2018sentence}, where we first fine-tune a pre-trained mBART model on our Non-Poetic data and then fine-tune the best model checkpoint on our Poetic data. The results for this experiment also did not lead to any significant difference in performance.

The third insight is that language-family-specific multilingual fine-tuning (Poetic LangFamily) helps in some of the languages when compared to multilingual fine-tuning on all languages (Poetic All). We also ran a preliminary experiment where we tested if multilingual fine-tuning with a dissimilar language hurts the performance compared to  fine-tuning with a language from the same language family (e.g., De and It vs. De and Nl). Our initial experiments show that fine-tuning on  languages from the same language family helps compared to  languages from different language family.

Last but not least, we notice that the multilingual fine-tuned model on poetic data (Poetic All) is consistently better than the bilingual fine-tuned model on poetic data (Poetic) across all languages.

\begin{table}
\small
\centering
\begin{tabular}{|l|c|c|}
\hline
        & \textbf{NonPoetic Best}              &  \textbf{Poetic Best} \\ \hline
Ru-En & 20\%           & 80\%  \\ \hline
Es-En & 0\%            & 100\%     \\ \hline
Pt-En &   40\%      &  60\%          \\\hline
De-En &     28\%             & 72\%        \\\hline
It-En &       28\%       &   72\%   \\\hline
Nl-En &         0\%     &    100\%  \\\hline
\end{tabular}
\caption{\label{human} Human evaluation in terms of preference between \textit{multilingual} fine-tuning on Non-Poetic data vs Poetic data, in terms of \textit{faithfulness}. Significant difference $( \alpha < 0.05)$ via Wilcoxon signed-rank test.}
\vspace{-2ex}
\end{table}

While we show that multilingual fine-tuning is an effective way to improve performance on low resource poetic data, we believe techniques like \textit{iterative backtranslation} \cite{hoang-etal-2018-iterative} with sophisticated techniques for data selection \cite{dou-etal-2020-dynamic} or domain repair \cite{wei-etal-2020-iterative} could improve the performance of model trained on Poetic data. We leave this for future work. 

\paragraph{Zero-Shot Performance on Unseen Languages}
We test the generalization capabilities of our model fine-tuned on poetic data using poetry written in languages not seen during fine-tuning. We compare the zero-shot performance of our model fine-tuned multilingually on poetic data (excluding  the unseen languages) to the Non-Poetic Multi (OPUS) and Non-Poetic  Multi  (ML50) model. We chose Ukrainian, Romanian and Swedish poetry given the fact that our model is fine-tuned on poetry belonging to languages from the Slavic, Romance, and Germanic families. Table \ref{table:zs} shows that our multilingually fine-tuned poetic model outperforms the other two multilingual models fine-tuned on Non-Poetic data, even though the languages were not contained in the fine-tuning data. This suggests that performance improvements of poetic fine-tuning are not only due to language-specific training data, but rather to multilinguality, presence of language family related data, as well as poetic style. These corroborate recent findings by \citet{ko-etal-2021-adapting} who adapt high-resource NMT models to translate low-resource related languages without parallel data. They exploit the fact that some low-resource languages are linguistically related or similar to high-resource languages, and often share many lexical or syntactic features.

\begin{table}
\centering
\small
\begin{tabular}{|l|l|l|l|l|}
\hline
                          &    & BLEU & BERTScore & COMET \\ \hline
\multirow{3}{*}{Ukranian} & M1 & 9.2  & 64.2   &  -39.61 \\ \cline{2-5}
                          & M2 & 9.1  & 65.0    & -40.46 \\ \cline{2-5}
                          & M3 & \textbf{15.1} & \textbf{67.3}   & \textbf{-32.10} \\ \hline\hline
                          
\multirow{3}{*}{Romanian} & M1 & \textbf{30.1} & 74.7  &  13.71  \\ \cline{2-5} 
                          & M2 & 24.4 & 73.6  & 9.43   \\ \cline{2-5} 
                          & M3 & 29.9 & \textbf{76.1}   &  \textbf{18.15} \\ \hline\hline
                          
\multirow{3}{*}{Swedish} & M1 & 14.3 & 68.0    & -24.21\\ \cline{2-5} 
                        & M2 & 16.6 & 66.4    & -30.47 \\ \cline{2-5} 
                          & M3 & \textbf{19.5} & \textbf{71.3}    & \textbf{-14.97}  \\ \hline                          
\end{tabular}
\caption{\label{table:zs} Zero-shot experiments. M1=Non-Poetic Multi(ML50); M2=Non-Poetic Multi(OPUS); M3=Poetic All.
Significant $( \alpha < 0.005)$ via Wilcoxon signed-rank test}
\end{table}

\section{Shortcomings of Style Transfer Techniques as a Post-Editing tool}
We evaluate whether style transfer techniques could help attenuate the shortcomings of translation models trained on non-poetic data. We use the romantic poetry style transfer model provided by \citet{style20} to paraphrase our non-poetic translations. This is the only available poetic style transfer model to our knowledge.To control for faithfulness, we generate $20$ outputs for each input (i.e., non-poetic translations) using nucleus sampling  ($p=0.6$), we then select the sentence that has the highest similarity score with input using the SIM model by \citet{wieting-etal-2019-beyond}.

The style transfer experiments decrease performance across all languages on both BLEU and BERTScore metrics as evaluated on the Multi(OPUS) model (see results in Table \ref{table:decrease}).

Qualitatively, this may happen due to errors cascading from incorrect translations by the non-poetic model, introduction of archaic language where it is not appropriate, and change in meaning. An example output is provided in Table \ref{table:badtransf}.  
\begin{table}
\small
\centering
\begin{tabular}{|l|l|l|}
\hline
            & \textbf{BLEU} & \textbf{BERTScore} \\ \hline
\textbf{RU} & 5.75 ({\color{red}-7.07})  & 59.91 ({\color{red}-7.29})      \\ \hline
\textbf{ES} & 6.42 ({\color{red}-21.58}) & 62.11 ({\color{red}-13.49})     \\ \hline
\textbf{PT} & 5.11 ({\color{red}-4.09})  & 58.24  ({\color{red}-5.76})     \\ \hline
\textbf{IT} & 6.13 ({\color{red}-16.77}) & 58.72 ({\color{red}-12.38})     \\ \hline
\textbf{DE} &   5.98 ({\color{red}-11.82}) & 60.07 ({\color{red}-10.83})                 \\ \hline
\textbf{NL} & 6.96 ({\color{red}-19.14}) & 60.95  ({\color{red}-11.95})    \\ \hline
\end{tabular}
\caption{BLEU and BERTScore after style transfer applied to the Multi(OPUS) configuration. Value in parenthesis reports decrease from the score obtained just by using Multi(OPUS).}
\label{table:decrease}
\end{table}

\begin{table}[h]
\small
\centering
\begin{tabular}{|l|l|}
\hline
\textbf{Gold} & What fun it is, with feet in sharp steel shod, \\ \hline
\textbf{M}    & How fun it is to wear iron-clad shoes,         \\ \hline
\textbf{M+ST} & Their iron shoes are saucy fun,                \\ \hline
\end{tabular}
\caption{Style transfer example. M=Multi(OPUS)}
\label{table:badtransf}
\end{table}

\section{Analysis}
\begin{figure}[t]
    \centering
    \includegraphics[width=0.45
    \textwidth]{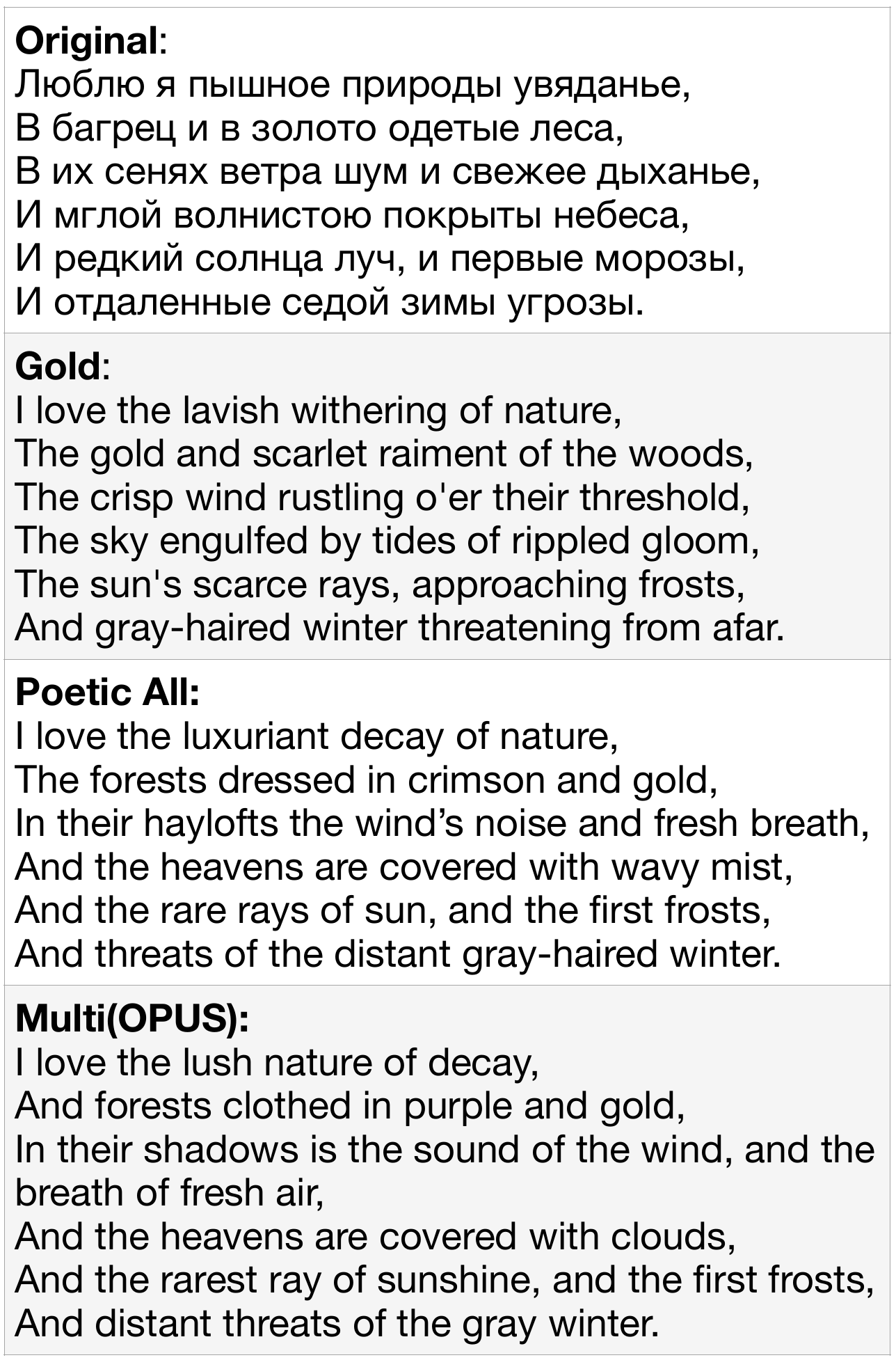}
    \caption{Example Russian-English translation}
    \label{fig:ex_ru}
\end{figure}

\begin{figure}[t]
    \centering
    \includegraphics[width=0.48
    \textwidth]{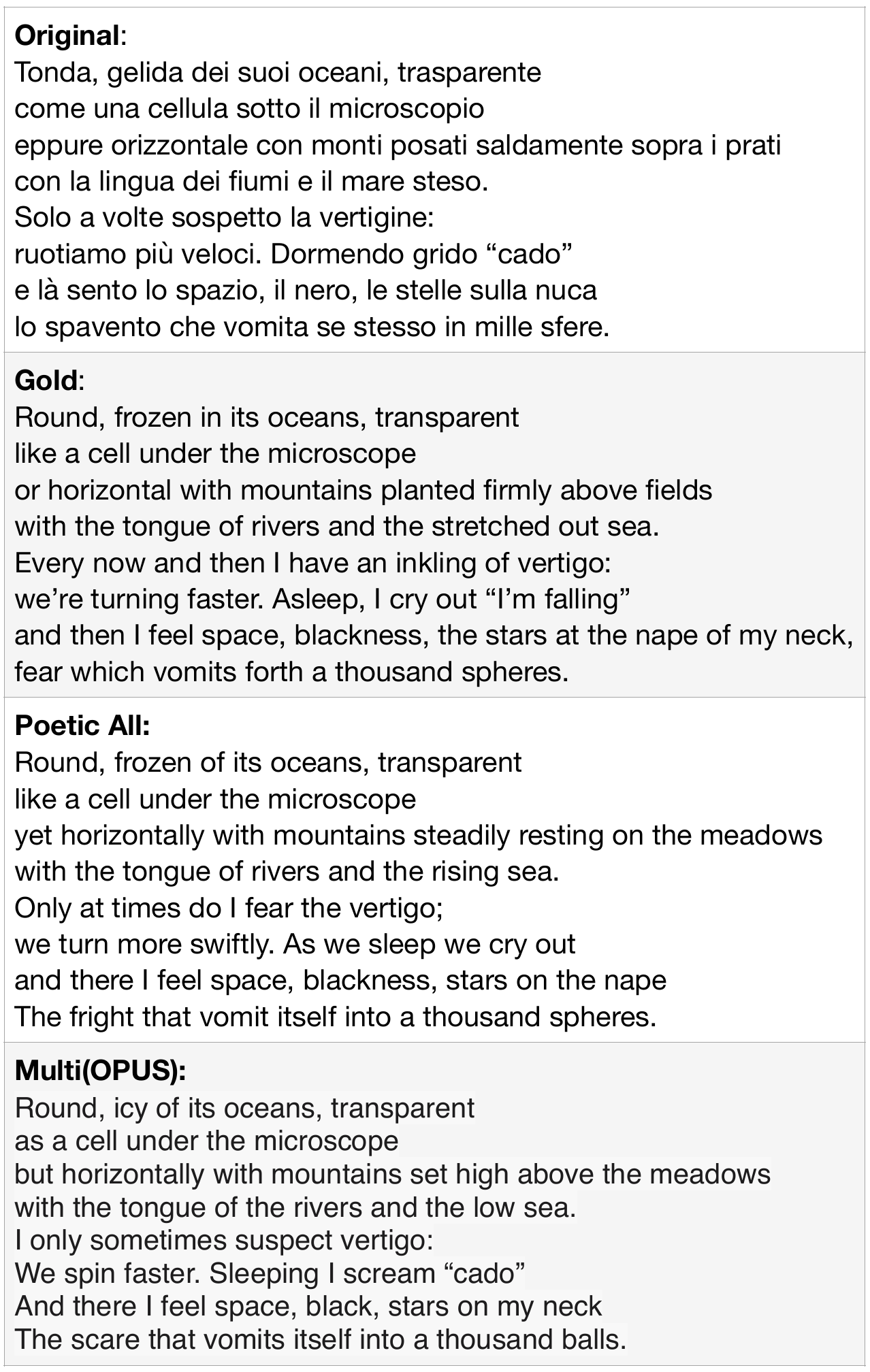}
    \caption{Example Italian-English translation}
    \label{fig:ex_it}
\end{figure}

\begin{figure}[t]
    \centering
    \includegraphics[width=0.45
    \textwidth]{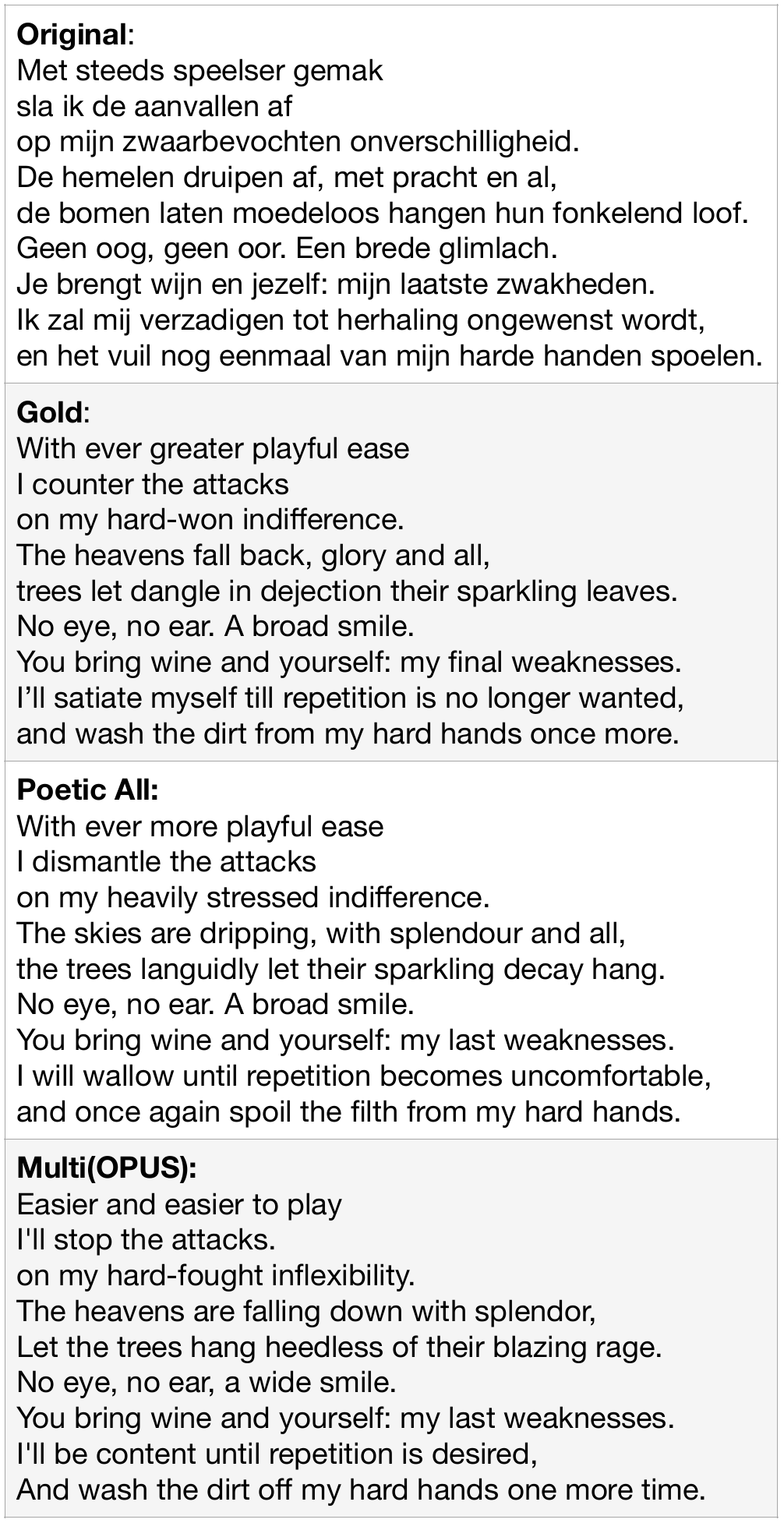}
    \caption{Example Dutch-English translation}
    \label{fig:ex_nl}
\end{figure}

It is well-known that occasionally NMT systems have a tendency 
to generate translations that are grammatically correct but unrelated to the source sentence particularly for low-resource settings (e.g., hallucinate words that are not mentioned in the source language) \cite{arthur-etal-2016-incorporating,koehn-knowles-2017-six}.
Pre-trained multilingual language models and techniques like multilingual training or fine-tuning can indeed be effective for dealing with low-resource data such as poetry as seen in Figures \ref{fig:ex_ru}, \ref{fig:ex_it}, \ref{fig:ex_nl}, showing examples of poetic translations by Poetic All and Multi(OPUS) configurations.
However, it is surprising that even a model trained on 6M parallel lines from OPUS(100) performs worse than models trained on in-domain data that is 35X smaller.

\vspace{0.1pt}

\begin{table}
\centering
\small
\begin{tabular}{|l|l|l|}
\hline
Gold      & \multicolumn{2}{l|}{\begin{tabular}[c]{@{}l@{}} \textit{Of a temple rising up in the gloom.}\end{tabular}}                  \\ \hline
PoeticAll    & \multicolumn{2}{l|}{\begin{tabular}[c]{@{}l@{}}\textit{Of a temple that rises in the dark,}\end{tabular}}                       \\ \hline
NonPoetic & \multicolumn{2}{l|}{\begin{tabular}[c]{@{}l@{}} \textit{A temple rising in the twinkling of an eye}\end{tabular}} \\ \hline\hline
Gold      & \multicolumn{2}{l|}{\begin{tabular}[c]{@{}l@{}} \textit{stand sails of smoke}\end{tabular}}                  \\ \hline
PoeticAll    & \multicolumn{2}{l|}{\begin{tabular}[c]{@{}l@{}}\textit{stand the sails of the smoke}\end{tabular}}                       \\ \hline
NonPoetic & \multicolumn{2}{l|}{\begin{tabular}[c]{@{}l@{}} \textit{There are sails of smoke}\end{tabular}}\\ \hline
\end{tabular}
\caption{\label{metaphor}Examples where metaphoric expressions are lost when translated using model fine-tuned multilingually on Non-Poetic (OPUS) data.}
\end{table}

Table \ref{metaphor} shows how model fine-tuned multilingually on non-poetic data suffer from loss of metaphoric expression in poetry, while a model fine-tuned multilingually 
on Poetic data is able to capture it. Table \ref{hallucination} shows how every model except our best poetic model fine-tuned multilingually suffer from hallucinations. The Non-Poetic model, while fluent to the reader, is not faithful to the original translation.

\section{Related work}
\paragraph{Domain adaptation in neural machine translation}
\citet{chu-wang-2018-survey} categorize domain adaptation for NMT in two groups: data centric and model centric. Data  centric techniques mostly focus on data augmentation for limited parallel  corpora of low-resource languages. For example, \citet{currey-etal-2017-copied} propose copying the target data to the source side to incorporate monolingual training data for low-resource languages. Back-translation has been  used for synthetic parallel corpora generation \cite{sennrich-etal-2016-improving}. To improve performance on specific domains, \citet{chu2017empirical} augment corpora with tags to indicate specific domains. A conventional model-centric approach is fine-tuning on in-domain parallel corpora or on mixed in-domain and out-of-domain corpora \cite{chu-wang-2018-survey}. In our work, we deal with a model-centric approach where we leverage a multilingual pre-trained model (mBART) and then fine-tune it multilingually on in-domain corpus. 
Recently, \citet{hu-etal-2019-domain} introduced a domain adaptation technique using lexicon induction, where large amounts of monolingual data are leveraged to find translations of in-domain unseen words. However, word-level lexicon induction might not be the most useful augmentation technique in our case, since poetic text deals with multi-word unseen phenomena such as metaphors.

\begin{table}
\small
\begin{tabular}{|p{1.2cm}|p{5.45cm}|}
\hline
Russian                                                    & \begin{tabular}[c]{@{}l@{}}\begin{otherlanguage*}{russian}Медуницы и осы тяжелую розу сосут.
\end{otherlanguage*}\\ \begin{otherlanguage*}{russian}Человек умирает.
\end{otherlanguage*}\\ \begin{otherlanguage*}{russian}Песок остывает согретый,\end{otherlanguage*}\end{tabular}       \\ \hline
NonPoetic                                                  & \begin{tabular}[c]{@{}l@{}}\textit{The \textbf{\color{red}vines and the leaves are} the heavy roses.}\\ \textit{A man dies. The sand is \textbf{\color{red}boiled down},}\end{tabular} \\ \hline
Poetic                                               & \begin{tabular}[c]{@{}l@{}}\textit{Honeycombs and wasps suck the heavy rose.}\\ \textit{Man dies. The warm sand cools,}\end{tabular}           \\ \hline
Gold                                                       & \begin{tabular}[c]{@{}l@{}}\textit{Bees and wasps suck the heavy rose.}\\ \textit{Man dies. The heated sand cools,}\end{tabular}               \\ \hline
\begin{tabular}[c]{@{}l@{}}Google\\ Translate\end{tabular} & \begin{tabular}[c]{@{}l@{}}\textit{\textbf{\color{red}Lungs} and wasps suck a heavy rose.}\\ \textit{The man is dying. Warmed sand is cooling}\end{tabular}        \\ \hline
\end{tabular}
\caption{\label{hallucination} Table showing hallucinations by other NonPoetic models including Google Translate, the ubiquitous translation behemoth.}
\end{table}

\paragraph{Poetic and literary translation}
\citet{jones-irvine-2013-un} discuss the difficulties of faithful machine translation of literary text in terms of the competing objectives of staying faithful to the original text but, on the other hand, trying to convey the experience of reading a literary piece to the reader. \citet{besacier-schwartz-2015-automated} conduct a pilot study of how suitable an MT+PE (machine translation + post-editing) pipeline would be for literary translation, concluding that their SMT approach could produce ``acceptable and rather readable'' translations.
\citet{matusov-2019-challenges} found that adapting NMT systems to literary content leads to improved automatic evaluation metrics on literary prose as compared to general domain NMT systems. \citet{kuzman-etal-2019-neural} found that Goolge NMT outperformed bespoke NMT models tailored to literature for English-Slovene literary translations. \citet{Toral2020MachineTO} perform a comprehensive human and automatic evaluation  (using BLEU) of NMT using Transformers \cite{NIPS2017_3f5ee243} for English-Catalan translation of novels, finding that domain-specific models lead to performance improvements judging by all evaluation techniques. \citet{fonteyne-etal-2020-literary} conduct a document-level evaluation of the translation of Agatha Christie's novel from English to Dutch using
Goolge's NMT system and found that most frequent issues were incorrect translation, coherence, style, and register.

Even though a lot of work has been done in the direction of automatic literary translation, automatic poetry translation is still in its infancy. \citet{genzel-etal-2010-poetic} produce poetry translations with meter and rhyme using phrase-based statistical MT approaches. \citet{ghazvininejad2018neural} present a neural poetry translation system that focuses on form rather than meaning. They also only focus on poetry translation from French to English, and their code or data is not publicly available. In our work, the focus is on faithfulness and the ability to preserve figurative language in translation across multiple languages.

\section{Conclusion and Future Work}
We release poetic parallel corpora for $6$ language pairs. Our work shows the clear benefit of domain adaptation for poetry translation. It further shows that improvements can be achieved by leveraging multilingual fine-tuning, and that the improvements transfer to unseen languages. Future directions include addition of new languages and larger corpora, adapting low-resource machine translation techniques for poetry translation, translating to languages that are morphologically richer than English, as well as working on better evaluation metrics to detect hallucinations. While computational methods for poetry translation may never outperform the human standard, we hope our contributions spark interest in the machine translation community to take up this rather challenging task. Additionally, by open-sourcing our work we hope to provide a helpful resource for professional translators.

\section*{Ethical Considerations}
Although we use language models trained on data collected from the Web, which have been shown to have issues with gender bias and abusive language  \cite{sheng-etal-2019-woman, wallace-etal-2019-universal} even in the multilingual space \cite{zhao-etal-2020-gender}, the inductive bias of our models should limit inadvertent negative impacts. Unlike model variants such as GPT, mBART is a conditional language model, which provides more control of the generated output. Our poetic parallel corpora are unlikely to contain toxic text and underwent manual inspection by the authors.

Technological advances in machine translation have had both positive and negative effects. Translation technology can diminish translators' professional autonomy as well as endanger professional translators’ livelihood. Moreover, given the fact that most people resort to models trained on non-literary text for literary translation could harms us in many ways. One such example is the negative influence caused by ungrammatical or unidiomatic language on readers’ linguistic skills in the target language, especially in the case of child readers. The low quality of literary translations can prevent the transfer of literary ideas and repertoires from one culture to another. Our work on poetry translation with a focus on faithfulness tries to bridge that gap. We believe interactive, human-in-the-loop MT systems designed especially for literary or poetic translation such as ours might speed up literary translators’ work and make it more enjoyable. 

Like \cite{petrilli2014sign} we believe too that automatic translation will not lead to the exclusion of human translators. Rather, it will increase human-machine interaction and continue enhancing human performance.  
Finally, we want to acknowledge all human translators who posted their work open-sourced on the websites we collected the data from. For our train and validation splits, the poems were broken down line by line and shuffled randomly. They do not contain any metadata and as such cannot reproduce the creative value of the original poems.

\bibliography{anthology,custom}

\begin{thebibliography}{55}
\expandafter\ifx\csname natexlab\endcsname\relax\def\natexlab#1{#1}\fi

\bibitem[{Aharoni et~al.(2019)Aharoni, Johnson, and
  Firat}]{aharoni-etal-2019-massively}
Roee Aharoni, Melvin Johnson, and Orhan Firat. 2019.
\newblock \href {https://doi.org/10.18653/v1/N19-1388} {Massively multilingual
  neural machine translation}.
\newblock In \emph{Proceedings of the 2019 Conference of the North {A}merican
  Chapter of the Association for Computational Linguistics: Human Language
  Technologies, Volume 1 (Long and Short Papers)}, pages 3874--3884,
  Minneapolis, Minnesota. Association for Computational Linguistics.

\bibitem[{Arthur et~al.(2016)Arthur, Neubig, and
  Nakamura}]{arthur-etal-2016-incorporating}
Philip Arthur, Graham Neubig, and Satoshi Nakamura. 2016.
\newblock \href {https://doi.org/10.18653/v1/D16-1162} {Incorporating discrete
  translation lexicons into neural machine translation}.
\newblock In \emph{Proceedings of the 2016 Conference on Empirical Methods in
  Natural Language Processing}, pages 1557--1567, Austin, Texas. Association
  for Computational Linguistics.

\bibitem[{Besacier and Schwartz(2015)}]{besacier-schwartz-2015-automated}
Laurent Besacier and Lane Schwartz. 2015.
\newblock \href {https://doi.org/10.3115/v1/W15-0713} {Automated translation of
  a literary work: A pilot study}.
\newblock In \emph{Proceedings of the Fourth Workshop on Computational
  Linguistics for Literature}, pages 114--122, Denver, Colorado, USA.
  Association for Computational Linguistics.

\bibitem[{Chen et~al.(2019)Chen, Yi, Sun, Li, Yang, and Guo}]{ijcai2019-684}
Huimin Chen, Xiaoyuan Yi, Maosong Sun, Wenhao Li, Cheng Yang, and Zhipeng Guo.
  2019.
\newblock \href {https://doi.org/10.24963/ijcai.2019/684}
  {Sentiment-controllable chinese poetry generation}.
\newblock In \emph{Proceedings of the Twenty-Eighth International Joint
  Conference on Artificial Intelligence, {IJCAI-19}}, pages 4925--4931.
  International Joint Conferences on Artificial Intelligence Organization.

\bibitem[{Chu et~al.(2017)Chu, Dabre, and Kurohashi}]{chu2017empirical}
Chenhui Chu, Raj Dabre, and Sadao Kurohashi. 2017.
\newblock \href {http://arxiv.org/abs/1701.03214} {An empirical comparison of
  simple domain adaptation methods for neural machine translation}.

\bibitem[{Chu and Wang(2018)}]{chu-wang-2018-survey}
Chenhui Chu and Rui Wang. 2018.
\newblock \href {https://www.aclweb.org/anthology/C18-1111} {A survey of domain
  adaptation for neural machine translation}.
\newblock In \emph{Proceedings of the 27th International Conference on
  Computational Linguistics}, pages 1304--1319, Santa Fe, New Mexico, USA.
  Association for Computational Linguistics.

\bibitem[{Currey et~al.(2017)Currey, Miceli~Barone, and
  Heafield}]{currey-etal-2017-copied}
Anna Currey, Antonio~Valerio Miceli~Barone, and Kenneth Heafield. 2017.
\newblock \href {https://doi.org/10.18653/v1/W17-4715} {Copied monolingual data
  improves low-resource neural machine translation}.
\newblock In \emph{Proceedings of the Second Conference on Machine
  Translation}, pages 148--156, Copenhagen, Denmark. Association for
  Computational Linguistics.

\bibitem[{Deng et~al.(2019)Deng, Wang, Liang, Chen, Xie, Zhuang, Wang, and
  Xiao}]{deng2019iterative}
Liming Deng, Jie Wang, Hangming Liang, Hui Chen, Zhiqiang Xie, Bojin Zhuang,
  Shaojun Wang, and Jing Xiao. 2019.
\newblock \href {http://arxiv.org/abs/1911.13182} {An iterative polishing
  framework based on quality aware masked language model for chinese poetry
  generation}.

\bibitem[{Dou et~al.(2020)Dou, Anastasopoulos, and
  Neubig}]{dou-etal-2020-dynamic}
Zi-Yi Dou, Antonios Anastasopoulos, and Graham Neubig. 2020.
\newblock \href {https://doi.org/10.18653/v1/2020.emnlp-main.475} {Dynamic data
  selection and weighting for iterative back-translation}.
\newblock In \emph{Proceedings of the 2020 Conference on Empirical Methods in
  Natural Language Processing (EMNLP)}, pages 5894--5904, Online. Association
  for Computational Linguistics.

\bibitem[{Fonteyne et~al.(2020)Fonteyne, Tezcan, and
  Macken}]{fonteyne-etal-2020-literary}
Margot Fonteyne, Arda Tezcan, and Lieve Macken. 2020.
\newblock \href {https://www.aclweb.org/anthology/2020.lrec-1.468} {Literary
  machine translation under the magnifying glass: Assessing the quality of an
  {NMT}-translated detective novel on document level}.
\newblock In \emph{Proceedings of the 12th Language Resources and Evaluation
  Conference}, pages 3790--3798, Marseille, France. European Language Resources
  Association.

\bibitem[{Frost(1961)}]{frost1961conversations}
Robert Frost. 1961.
\newblock \emph{Conversations on the Craft of Poetry}.
\newblock Holt, Rinehart and Winston.

\bibitem[{Genzel et~al.(2010)Genzel, Uszkoreit, and
  Och}]{genzel-etal-2010-poetic}
Dmitriy Genzel, Jakob Uszkoreit, and Franz Och. 2010.
\newblock \href {https://aclanthology.org/D10-1016} {{``}poetic{''} statistical
  machine translation: Rhyme and meter}.
\newblock In \emph{Proceedings of the 2010 Conference on Empirical Methods in
  Natural Language Processing}, pages 158--166, Cambridge, MA. Association for
  Computational Linguistics.

\bibitem[{Ghazvininejad et~al.(2018)Ghazvininejad, Choi, and
  Knight}]{ghazvininejad2018neural}
Marjan Ghazvininejad, Yejin Choi, and Kevin Knight. 2018.
\newblock Neural poetry translation.
\newblock In \emph{Proceedings of the 2018 Conference of the North American
  Chapter of the Association for Computational Linguistics: Human Language
  Technologies, Volume 2 (Short Papers)}, pages 67--71.

\bibitem[{Ghazvininejad et~al.(2016)Ghazvininejad, Shi, Choi, and
  Knight}]{ghazvininejad2016generating}
Marjan Ghazvininejad, Xing Shi, Yejin Choi, and Kevin Knight. 2016.
\newblock Generating topical poetry.
\newblock In \emph{Proceedings of the 2016 Conference on Empirical Methods in
  Natural Language Processing}, pages 1183--1191.

\bibitem[{Graham et~al.(2013)Graham, Baldwin, Moffat, and
  Zobel}]{graham-etal-2013-continuous}
Yvette Graham, Timothy Baldwin, Alistair Moffat, and Justin Zobel. 2013.
\newblock \href {https://aclanthology.org/W13-2305} {Continuous measurement
  scales in human evaluation of machine translation}.
\newblock In \emph{Proceedings of the 7th Linguistic Annotation Workshop and
  Interoperability with Discourse}, pages 33--41, Sofia, Bulgaria. Association
  for Computational Linguistics.

\bibitem[{Gururangan et~al.(2020)Gururangan, Marasovi{\'c}, Swayamdipta, Lo,
  Beltagy, Downey, and Smith}]{gururangan-etal-2020-dont}
Suchin Gururangan, Ana Marasovi{\'c}, Swabha Swayamdipta, Kyle Lo, Iz~Beltagy,
  Doug Downey, and Noah~A. Smith. 2020.
\newblock \href {https://doi.org/10.18653/v1/2020.acl-main.740} {Don{'}t stop
  pretraining: Adapt language models to domains and tasks}.
\newblock In \emph{Proceedings of the 58th Annual Meeting of the Association
  for Computational Linguistics}, pages 8342--8360, Online. Association for
  Computational Linguistics.

\bibitem[{H{\"a}m{\"a}l{\"a}inen and
  Alnajjar(2019)}]{hamalainen-alnajjar-2019-generating}
Mika H{\"a}m{\"a}l{\"a}inen and Khalid Alnajjar. 2019.
\newblock \href {https://doi.org/10.18653/v1/D19-1617} {Generating modern
  poetry automatically in {F}innish}.
\newblock In \emph{Proceedings of the 2019 Conference on Empirical Methods in
  Natural Language Processing and the 9th International Joint Conference on
  Natural Language Processing (EMNLP-IJCNLP)}, pages 5999--6004, Hong Kong,
  China. Association for Computational Linguistics.

\bibitem[{He et~al.(2020)He, Liu, Gao, and Chen}]{he2020deberta}
Pengcheng He, Xiaodong Liu, Jianfeng Gao, and Weizhu Chen. 2020.
\newblock Deberta: Decoding-enhanced bert with disentangled attention.
\newblock \emph{arXiv preprint arXiv:2006.03654}.

\bibitem[{Hoang et~al.(2018)Hoang, Koehn, Haffari, and
  Cohn}]{hoang-etal-2018-iterative}
Vu~Cong~Duy Hoang, Philipp Koehn, Gholamreza Haffari, and Trevor Cohn. 2018.
\newblock \href {https://doi.org/10.18653/v1/W18-2703} {Iterative
  back-translation for neural machine translation}.
\newblock In \emph{Proceedings of the 2nd Workshop on Neural Machine
  Translation and Generation}, pages 18--24, Melbourne, Australia. Association
  for Computational Linguistics.

\bibitem[{Hopkins and Kiela(2017)}]{hopkins-kiela-2017-automatically}
Jack Hopkins and Douwe Kiela. 2017.
\newblock \href {https://doi.org/10.18653/v1/P17-1016} {Automatically
  generating rhythmic verse with neural networks}.
\newblock In \emph{Proceedings of the 55th Annual Meeting of the Association
  for Computational Linguistics (Volume 1: Long Papers)}, pages 168--178,
  Vancouver, Canada. Association for Computational Linguistics.

\bibitem[{Hu et~al.(2019)Hu, Xia, Neubig, and Carbonell}]{hu-etal-2019-domain}
Junjie Hu, Mengzhou Xia, Graham Neubig, and Jaime Carbonell. 2019.
\newblock \href {https://doi.org/10.18653/v1/P19-1286} {Domain adaptation of
  neural machine translation by lexicon induction}.
\newblock In \emph{Proceedings of the 57th Annual Meeting of the Association
  for Computational Linguistics}, pages 2989--3001, Florence, Italy.
  Association for Computational Linguistics.

\bibitem[{Jones(2011)}]{jones2011translation}
Francis~R Jones. 2011.
\newblock The translation of poetry.
\newblock In \emph{The Oxford handbook of translation studies}. Oxford
  University Press.

\bibitem[{Jones and Irvine(2013)}]{jones-irvine-2013-un}
Ruth Jones and Ann Irvine. 2013.
\newblock \href {https://www.aclweb.org/anthology/W13-2713} {The (un)faithful
  machine translator}.
\newblock In \emph{Proceedings of the 7th Workshop on Language Technology for
  Cultural Heritage, Social Sciences, and Humanities}, pages 96--101, Sofia,
  Bulgaria. Association for Computational Linguistics.

\bibitem[{Ko et~al.(2021)Ko, El-Kishky, Renduchintala, Chaudhary, Goyal,
  Guzm{\'a}n, Fung, Koehn, and Diab}]{ko-etal-2021-adapting}
Wei-Jen Ko, Ahmed El-Kishky, Adithya Renduchintala, Vishrav Chaudhary, Naman
  Goyal, Francisco Guzm{\'a}n, Pascale Fung, Philipp Koehn, and Mona Diab.
  2021.
\newblock \href {https://doi.org/10.18653/v1/2021.acl-long.66} {Adapting
  high-resource {NMT} models to translate low-resource related languages
  without parallel data}.
\newblock In \emph{Proceedings of the 59th Annual Meeting of the Association
  for Computational Linguistics and the 11th International Joint Conference on
  Natural Language Processing (Volume 1: Long Papers)}, pages 802--812, Online.
  Association for Computational Linguistics.

\bibitem[{Kocmi et~al.(2021)Kocmi, Federmann, Grundkiewicz, Junczys-Dowmunt,
  Matsushita, and Menezes}]{kocmi2021ship}
Tom Kocmi, Christian Federmann, Roman Grundkiewicz, Marcin Junczys-Dowmunt,
  Hitokazu Matsushita, and Arul Menezes. 2021.
\newblock To ship or not to ship: An extensive evaluation of automatic metrics
  for machine translation.
\newblock \emph{arXiv preprint arXiv:2107.10821}.

\bibitem[{Koehn and Knowles(2017)}]{koehn-knowles-2017-six}
Philipp Koehn and Rebecca Knowles. 2017.
\newblock \href {https://doi.org/10.18653/v1/W17-3204} {Six challenges for
  neural machine translation}.
\newblock In \emph{Proceedings of the First Workshop on Neural Machine
  Translation}, pages 28--39, Vancouver. Association for Computational
  Linguistics.

\bibitem[{Krishna et~al.(2020)Krishna, Wieting, and Iyyer}]{style20}
Kalpesh Krishna, John Wieting, and Mohit Iyyer. 2020.
\newblock Reformulating unsupervised style transfer as paraphrase generation.
\newblock In \emph{Empirical Methods in Natural Language Processing}.

\bibitem[{Kuzman et~al.(2019)Kuzman, Vintar, and
  Ar{\v{c}}an}]{kuzman-etal-2019-neural}
Taja Kuzman, {\v{S}}pela Vintar, and Mihael Ar{\v{c}}an. 2019.
\newblock \href {https://www.aclweb.org/anthology/W19-7301} {Neural machine
  translation of literary texts from {E}nglish to {S}lovene}.
\newblock In \emph{Proceedings of the Qualities of Literary Machine
  Translation}, pages 1--9, Dublin, Ireland. European Association for Machine
  Translation.

\bibitem[{Lewis et~al.(2019)Lewis, Liu, Goyal, Ghazvininejad, Mohamed, Levy,
  Stoyanov, and Zettlemoyer}]{lewis2019bart}
Mike Lewis, Yinhan Liu, Naman Goyal, Marjan Ghazvininejad, Abdelrahman Mohamed,
  Omer Levy, Ves Stoyanov, and Luke Zettlemoyer. 2019.
\newblock Bart: Denoising sequence-to-sequence pre-training for natural
  language generation, translation, and comprehension.
\newblock \emph{arXiv preprint arXiv:1910.13461}.

\bibitem[{Li et~al.(2020)Li, Zhang, Liu, and Shi}]{li-etal-2020-rigid}
Piji Li, Haisong Zhang, Xiaojiang Liu, and Shuming Shi. 2020.
\newblock \href {https://doi.org/10.18653/v1/2020.acl-main.68} {Rigid formats
  controlled text generation}.
\newblock In \emph{Proceedings of the 58th Annual Meeting of the Association
  for Computational Linguistics}, pages 742--751, Online. Association for
  Computational Linguistics.

\bibitem[{Liu et~al.(2020)Liu, Gu, Goyal, Li, Edunov, Ghazvininejad, Lewis, and
  Zettlemoyer}]{liu2020multilingual}
Yinhan Liu, Jiatao Gu, Naman Goyal, Xian Li, Sergey Edunov, Marjan
  Ghazvininejad, Mike Lewis, and Luke Zettlemoyer. 2020.
\newblock \href {https://doi.org/10.1162/tacl_a_00343} {Multilingual denoising
  pre-training for neural machine translation}.
\newblock \emph{Transactions of the Association for Computational Linguistics},
  8:726--742.

\bibitem[{Matusov(2019)}]{matusov-2019-challenges}
Evgeny Matusov. 2019.
\newblock \href {https://www.aclweb.org/anthology/W19-7302} {The challenges of
  using neural machine translation for literature}.
\newblock In \emph{Proceedings of the Qualities of Literary Machine
  Translation}, pages 10--19, Dublin, Ireland. European Association for Machine
  Translation.

\bibitem[{Papineni et~al.(2002)Papineni, Roukos, Ward, and Zhu}]{BLEU}
Kishore Papineni, Salim Roukos, Todd Ward, and Wei-Jing Zhu. 2002.
\newblock Bleu: a method for automatic evaluation of machine translation.
\newblock In \emph{Proceedings of the 40th annual meeting on association for
  computational linguistics}, pages 311--318. Association for Computational
  Linguistics.

\bibitem[{Petrilli(2014)}]{petrilli2014sign}
Susan Petrilli. 2014.
\newblock \emph{Sign studies and semioethics: Communication, translation and
  values}, volume~13.
\newblock Walter de Gruyter GmbH \& Co KG.

\bibitem[{Phang et~al.(2018)Phang, F{\'e}vry, and Bowman}]{phang2018sentence}
Jason Phang, Thibault F{\'e}vry, and Samuel~R Bowman. 2018.
\newblock Sentence encoders on stilts: Supplementary training on intermediate
  labeled-data tasks.
\newblock \emph{arXiv preprint arXiv:1811.01088}.

\bibitem[{Post(2018)}]{post-2018-call}
Matt Post. 2018.
\newblock \href {https://doi.org/10.18653/v1/W18-6319} {A call for clarity in
  reporting {BLEU} scores}.
\newblock In \emph{Proceedings of the Third Conference on Machine Translation:
  Research Papers}, pages 186--191, Brussels, Belgium. Association for
  Computational Linguistics.

\bibitem[{Raunak et~al.(2021)Raunak, Menezes, and
  Junczys-Dowmunt}]{raunak-etal-2021-curious}
Vikas Raunak, Arul Menezes, and Marcin Junczys-Dowmunt. 2021.
\newblock \href {https://doi.org/10.18653/v1/2021.naacl-main.92} {The curious
  case of hallucinations in neural machine translation}.
\newblock In \emph{Proceedings of the 2021 Conference of the North American
  Chapter of the Association for Computational Linguistics: Human Language
  Technologies}, pages 1172--1183, Online. Association for Computational
  Linguistics.

\bibitem[{Rei et~al.(2020)Rei, Stewart, Farinha, and
  Lavie}]{rei-etal-2020-comet}
Ricardo Rei, Craig Stewart, Ana~C Farinha, and Alon Lavie. 2020.
\newblock \href {https://doi.org/10.18653/v1/2020.emnlp-main.213} {{COMET}: A
  neural framework for {MT} evaluation}.
\newblock In \emph{Proceedings of the 2020 Conference on Empirical Methods in
  Natural Language Processing (EMNLP)}, pages 2685--2702, Online. Association
  for Computational Linguistics.

\bibitem[{Sennrich et~al.(2016)Sennrich, Haddow, and
  Birch}]{sennrich-etal-2016-improving}
Rico Sennrich, Barry Haddow, and Alexandra Birch. 2016.
\newblock \href {https://doi.org/10.18653/v1/P16-1009} {Improving neural
  machine translation models with monolingual data}.
\newblock In \emph{Proceedings of the 54th Annual Meeting of the Association
  for Computational Linguistics (Volume 1: Long Papers)}, pages 86--96, Berlin,
  Germany. Association for Computational Linguistics.

\bibitem[{Sheng et~al.(2019)Sheng, Chang, Natarajan, and
  Peng}]{sheng-etal-2019-woman}
Emily Sheng, Kai-Wei Chang, Premkumar Natarajan, and Nanyun Peng. 2019.
\newblock \href {https://doi.org/10.18653/v1/D19-1339} {The woman worked as a
  babysitter: On biases in language generation}.
\newblock In \emph{Proceedings of the 2019 Conference on Empirical Methods in
  Natural Language Processing and the 9th International Joint Conference on
  Natural Language Processing (EMNLP-IJCNLP)}, pages 3407--3412, Hong Kong,
  China. Association for Computational Linguistics.

\bibitem[{Tang et~al.(2020)Tang, Tran, Li, Chen, Goyal, Chaudhary, Gu, and
  Fan}]{tang2020multilingual}
Yuqing Tang, Chau Tran, Xian Li, Peng-Jen Chen, Naman Goyal, Vishrav Chaudhary,
  Jiatao Gu, and Angela Fan. 2020.
\newblock Multilingual translation with extensible multilingual pretraining and
  finetuning.
\newblock \emph{arXiv preprint arXiv:2008.00401}.

\bibitem[{Tiedemann(2012)}]{tiedemann-2012-parallel}
J{\"o}rg Tiedemann. 2012.
\newblock \href
  {http://www.lrec-conf.org/proceedings/lrec2012/pdf/463_Paper.pdf} {Parallel
  data, tools and interfaces in {OPUS}}.
\newblock In \emph{Proceedings of the Eighth International Conference on
  Language Resources and Evaluation ({LREC}'12)}, pages 2214--2218, Istanbul,
  Turkey. European Language Resources Association (ELRA).

\bibitem[{Toral et~al.(2020)Toral, Oliver, and Ballest'in}]{Toral2020MachineTO}
Antonio Toral, A.~Oliver, and Pau~Ribas Ballest'in. 2020.
\newblock Machine translation of novels in the age of transformer.
\newblock \emph{ArXiv}, abs/2011.14979.

\bibitem[{Uthus et~al.(2021)Uthus, Voitovich, and Mical}]{uthus2021augmenting}
David Uthus, Maria Voitovich, and RJ~Mical. 2021.
\newblock Augmenting poetry composition with verse by verse.
\newblock \emph{arXiv preprint arXiv:2103.17205}.

\bibitem[{Van~de Cruys(2020)}]{van2020automatic}
Tim Van~de Cruys. 2020.
\newblock Automatic poetry generation from prosaic text.
\newblock In \emph{Proceedings of the 58th Annual Meeting of the Association
  for Computational Linguistics}, pages 2471--2480.

\bibitem[{Vaswani et~al.(2017)Vaswani, Shazeer, Parmar, Uszkoreit, Jones,
  Gomez, Kaiser, and Polosukhin}]{NIPS2017_3f5ee243}
Ashish Vaswani, Noam Shazeer, Niki Parmar, Jakob Uszkoreit, Llion Jones,
  Aidan~N Gomez, \L~ukasz Kaiser, and Illia Polosukhin. 2017.
\newblock \href
  {https://proceedings.neurips.cc/paper/2017/file/3f5ee243547dee91fbd053c1c4a845aa-Paper.pdf}
  {Attention is all you need}.
\newblock In \emph{Advances in Neural Information Processing Systems},
  volume~30. Curran Associates, Inc.

\bibitem[{Wallace et~al.(2019)Wallace, Feng, Kandpal, Gardner, and
  Singh}]{wallace-etal-2019-universal}
Eric Wallace, Shi Feng, Nikhil Kandpal, Matt Gardner, and Sameer Singh. 2019.
\newblock \href {https://doi.org/10.18653/v1/D19-1221} {Universal adversarial
  triggers for attacking and analyzing {NLP}}.
\newblock In \emph{Proceedings of the 2019 Conference on Empirical Methods in
  Natural Language Processing and the 9th International Joint Conference on
  Natural Language Processing (EMNLP-IJCNLP)}, pages 2153--2162, Hong Kong,
  China. Association for Computational Linguistics.

\bibitem[{Wei et~al.(2020)Wei, Zhang, Chen, and Luo}]{wei-etal-2020-iterative}
Hao-Ran Wei, Zhirui Zhang, Boxing Chen, and Weihua Luo. 2020.
\newblock \href {https://doi.org/10.18653/v1/2020.emnlp-main.474} {Iterative
  domain-repaired back-translation}.
\newblock In \emph{Proceedings of the 2020 Conference on Empirical Methods in
  Natural Language Processing (EMNLP)}, pages 5884--5893, Online. Association
  for Computational Linguistics.

\bibitem[{Wieting et~al.(2019)Wieting, Berg-Kirkpatrick, Gimpel, and
  Neubig}]{wieting-etal-2019-beyond}
John Wieting, Taylor Berg-Kirkpatrick, Kevin Gimpel, and Graham Neubig. 2019.
\newblock \href {https://doi.org/10.18653/v1/P19-1427} {Beyond {BLEU}:training
  neural machine translation with semantic similarity}.
\newblock In \emph{Proceedings of the 57th Annual Meeting of the Association
  for Computational Linguistics}, pages 4344--4355, Florence, Italy.
  Association for Computational Linguistics.

\bibitem[{Williams et~al.(2017)Williams, Nangia, and
  Bowman}]{williams2017broad}
Adina Williams, Nikita Nangia, and Samuel~R Bowman. 2017.
\newblock A broad-coverage challenge corpus for sentence understanding through
  inference.
\newblock \emph{arXiv preprint arXiv:1704.05426}.

\bibitem[{Wolf et~al.(2020)Wolf, Debut, Sanh, Chaumond, Delangue, Moi, Cistac,
  Rault, Louf, Funtowicz, Davison, Shleifer, von Platen, Ma, Jernite, Plu, Xu,
  Le~Scao, Gugger, Drame, Lhoest, and Rush}]{wolf-etal-2020-transformers}
Thomas Wolf, Lysandre Debut, Victor Sanh, Julien Chaumond, Clement Delangue,
  Anthony Moi, Pierric Cistac, Tim Rault, Remi Louf, Morgan Funtowicz, Joe
  Davison, Sam Shleifer, Patrick von Platen, Clara Ma, Yacine Jernite, Julien
  Plu, Canwen Xu, Teven Le~Scao, Sylvain Gugger, Mariama Drame, Quentin Lhoest,
  and Alexander Rush. 2020.
\newblock \href {https://doi.org/10.18653/v1/2020.emnlp-demos.6} {Transformers:
  State-of-the-art natural language processing}.
\newblock In \emph{Proceedings of the 2020 Conference on Empirical Methods in
  Natural Language Processing: System Demonstrations}, pages 38--45, Online.
  Association for Computational Linguistics.

\bibitem[{Yang et~al.(2018)Yang, Sun, Yi, and Li}]{yang-etal-2018-stylistic}
Cheng Yang, Maosong Sun, Xiaoyuan Yi, and Wenhao Li. 2018.
\newblock \href {https://doi.org/10.18653/v1/D18-1430} {Stylistic {C}hinese
  poetry generation via unsupervised style disentanglement}.
\newblock In \emph{Proceedings of the 2018 Conference on Empirical Methods in
  Natural Language Processing}, pages 3960--3969, Brussels, Belgium.
  Association for Computational Linguistics.

\bibitem[{Yi et~al.(2020)Yi, Li, Yang, Li, and Sun}]{yi2020mixpoet}
Xiaoyuan Yi, Ruoyu Li, Cheng Yang, Wenhao Li, and Maosong Sun. 2020.
\newblock \href {http://arxiv.org/abs/2003.06094} {Mixpoet: Diverse poetry
  generation via learning controllable mixed latent space}.

\bibitem[{Zhang et~al.(2019)Zhang, Kishore, Wu, Weinberger, and
  Artzi}]{zhang2019bertscore}
Tianyi Zhang, Varsha Kishore, Felix Wu, Kilian~Q Weinberger, and Yoav Artzi.
  2019.
\newblock Bertscore: Evaluating text generation with bert.
\newblock \emph{arXiv preprint arXiv:1904.09675}.

\bibitem[{Zhao et~al.(2020)Zhao, Mukherjee, Hosseini, Chang, and
  Hassan~Awadallah}]{zhao-etal-2020-gender}
Jieyu Zhao, Subhabrata Mukherjee, Saghar Hosseini, Kai-Wei Chang, and Ahmed
  Hassan~Awadallah. 2020.
\newblock \href {https://doi.org/10.18653/v1/2020.acl-main.260} {Gender bias in
  multilingual embeddings and cross-lingual transfer}.
\newblock In \emph{Proceedings of the 58th Annual Meeting of the Association
  for Computational Linguistics}, pages 2896--2907, Online. Association for
  Computational Linguistics.

\end{thebibliography}
\bibliographystyle{acl_natbib}

\clearpage

\end{document}